\documentclass[sn-mathphys]{sn-jnl}%

\usepackage{graphicx}
\usepackage{xcolor}
\usepackage{booktabs}
\usepackage{array}
\usepackage{tabularx}
\usepackage{comment}

\jyear{2022}%

\theoremstyle{thmstyleone}%

\theoremstyle{thmstyletwo}%

\theoremstyle{thmstylethree}%

\begin{document}

\title[DisguisOR]{DisguisOR: Holistic Face Anonymization for the Operating Room}

\author{\fnm{Lennart} \sur{Bastian$^*$}}
\author{\fnm{Tony Danjun} \sur{Wang$^*$}}
\author{\fnm{Tobias} \sur{Czempiel}}
\author{\fnm{Benjamin} \sur{Busam}}
\author{\fnm{Nassir} \sur{Navab}}

\affil{\orgdiv{Computer Aided Medical Procedures}, \orgname{Technical University of Munich}, \orgaddress{\city{Munich},  \country{Germany}}}
\affil{\small{\\$^*$ Equal Contribution. \qquad Contact: \{first\}.\{last\}@tum.de}}

\abstract{
\textbf{Purpose:}
Recent advances in Surgical Data Science (SDS) have contributed to an increase in video recordings from hospital environments. While methods such as surgical workflow recognition show potential in increasing the quality of patient care, the quantity of video data has surpassed the scale at which images can be manually anonymized. 
Existing automated 2D anonymization methods under-perform in Operating Rooms (OR), due to occlusions and obstructions. 
We propose to anonymize multi-view OR recordings using 3D data from multiple camera streams.

\textbf{Methods:}
RGB and depth images from multiple cameras are fused into a 3D point cloud representation of the scene.
We then detect each individual's face in 3D by regressing a parametric human mesh model onto detected 3D human keypoints and aligning the face mesh with the fused 3D point cloud. 
The mesh model is rendered into every acquired camera view, replacing each individual's face.

\textbf{Results:}
Our method shows promise in locating faces at a higher rate than existing approaches. DisguisOR produces geometrically consistent anonymizations for each camera view, enabling more realistic anonymization that is less detrimental to downstream tasks. 

\textbf{Conclusion:}
Frequent obstructions and crowding in operating rooms leaves significant room for improvement for off-the-shelf anonymization methods.
DisguisOR addresses privacy on a scene level and has the potential to facilitate further research in SDS.
}

\keywords{Anonymization, Face Detection, Multi-view Operating Rooms, Surgical Data Science, Surgical Workflow Recognition}

\maketitle

\section{Introduction}
\label{sec:introduction}
The past years have seen an increase in video acquisitions in hospitals and surgical environments. 
In the field of surgical data science (SDS), the analysis of endoscopic and laparoscopic frames is already an established research direction~\cite{tecno}. 
It aims to build cognitive systems capable of understanding the procedural steps of an intervention, for example, recognizing and localizing surgical tools~\cite{ml_for_surg_phase_recog}.
Closely related to the endoscopic frames are videos from externally mounted cameras, capturing the surgical scene from an outside perspective~\cite{mvor}.
These rich information sources build the foundation for analyzing and optimizing the workflow, essential for developing context-aware intelligent systems, improving patient quality of care, and advancing anomaly detection. 
However, video recordings of surgeries are still considered problematic due to strict privacy regulations established to protect both patients and medical staff. 
As manually anonymizing video frames no longer becomes feasible at scale, it is imperative to develop automatic de-identification methods to advance future research and facilitate SDS dataset curation.

Surgical operating rooms are frequently crowded and packed with medical equipment. 
Cameras can only be mounted at particular positions, leading to perspectives not usually found in conventional datasets~\cite{widerface}.
This poses challenges even for advanced anonymization methods, as they tend to perform poorly under partial occlusions and obscure camera angles~\cite{issenhuth_face_detection_2018}.
A few methods address the specific challenges of OR anonymization~\cite{face_off,issenhuth_face_detection_2018} from individual cameras.
Recent works propose addressing the OR's unique challenges by combining multi-view RGB-D data (\textit{multi-view}) to compensate for missed information in surgical workflow recognition~\cite{know_your_sensors,schmidt2021multi,mvor,sharghi2020automatic}. 
The existence of such \textit{multi-view} OR recordings requires anonymizing all camera views, as a failed anonymization in a single view breaches the privacy of the entire scene.

\begin{figure}[t!]
    \includegraphics[width=\columnwidth]{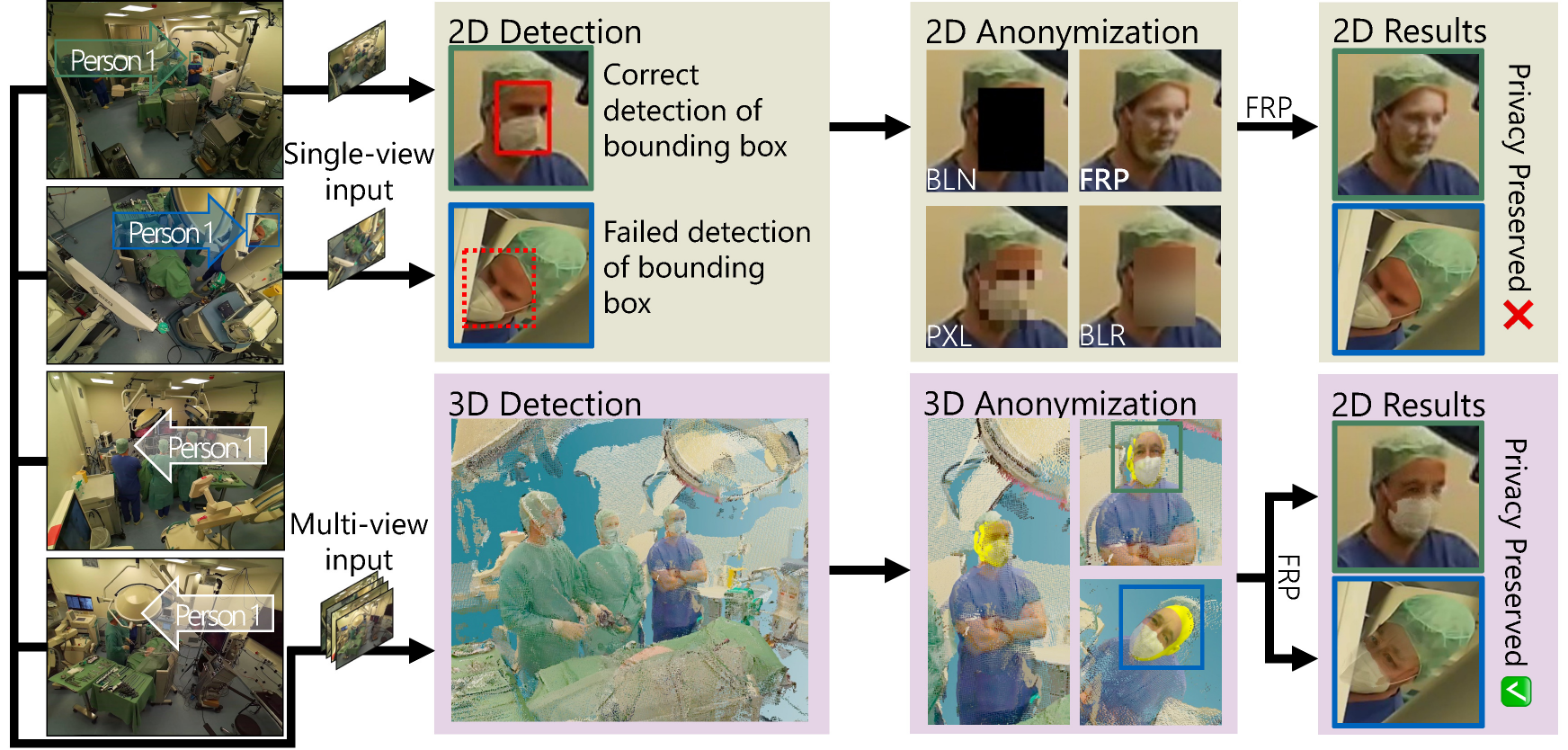}
    \caption{
        Holistic Face Anonymization From Scene Level Representation. Four views of a \textit{multi-view} OR acquisition are visualized on the left, highlighting person 1 in each view. Conventional methods detect and anonymize faces in each image individually (green boxes) through either blackening (BLN), pixelization (PXL), blurring (BLR), or face replacement (FRP)~\cite{face_off,mvor,deepprivacy}.
        Our framework (purple boxes) leverages \textit{multi-view} information to localize faces in 3D, enabling consistent anonymization in all four images
        }
    \label{fig:teaser_figure}
\end{figure}

We propose a novel anonymization approach for \textit{multi-view} recordings, which leverages 3D information to detect faces where conventional methods fail. 
We utilize a 3D mesh to accurately replace each detected person's face, preserving privacy as well as data integrity in all camera views.
In~\autoref{fig:teaser_figure}, we compare single and \textit{multi-view} approaches, highlighting the advantages of scene-level anonymization.
We additionally show that in comparison to existing methods, our face replacement yields images that harmonize well with the surgical environment as measured by image similarity.
Our main contributions can be summarized as follows:
\begin{itemize}
    \item We present a novel framework for accurate \textit{multi-view} 2D face localization by leveraging 3D information. 
    We further emphasize the necessity for consistent anonymization across all camera views using our proposed \textit{holistic recall}.
    \item We present a training-free, mesh-based anonymization method yielding complete control during the 3D face replacement step while generating more realistic results than existing state-of-the-art approaches.
    \item The images anonymized by our framework can be effectively utilized by downstream methods, as shown through experiments on image quality assessment and downstream face localization.
\end{itemize}

\section{Related Works}

\noindent\textbf{Face detection.} With the advent of public face detection benchmark datasets like WIDERFACE~\cite{widerface},
numerous deep learning-based face detectors were introduced in recent years~\cite{dsfd,tinaface,minaee2021going}. 
Such methods typically regress a bounding box onto the region where a face can be successfully identified in the image.
As WIDERFACE consists of annotated images from everyday scenarios, face detectors trained on this dataset can suffer in complex and crowded OR environments~\cite{issenhuth_face_detection_2018}. 
Occlusions and obstructions from medical equipment or personnel in close quarters, masks, and skull caps can lead to missed predictions and, ultimately, incomplete anonymizations.
While we also use 3D data for anonymization, our work diverges from 3D face recognition~\cite{zhou20183d}, where a scan of a 3D face is matched to a catalogue of face scans.

\noindent\textbf{Image anonymization.} Identity scrubbing can be achieved by removing the sensitive area, blurring, or pixelization~\cite{mvor}.
In the OR, standardized scrubs and gloves already obscure many possible landmarks, leaving the face as the primary identifier that could be used for re-identification, as previously established~\cite{issenhuth_face_detection_2018,face_off}.
A recent line of work has proposed to replace faces with artificially generated faces using GANs~\cite{cai2021generative,deepprivacy} or parametric face models~\cite{hybrid_model}. 
Such replacement methods tend to yield a more realistic-looking output, and the resulting anonymized area resembles the input more closely, which can positively affect downstream applications~\cite{cai2021generative}.
However, these methods typically contain a separate branch to handle face detection~\cite{deepprivacy} and thus suffer similarly in OR environments due to partial obstructions.

\noindent\textbf{Human Pose Estimation.} Using human pose estimation as an additional context to localize faces has been demonstrated as valuable~\cite{issenhuth_face_detection_2018}. The torso, shoulders, and arms provide useful cues for localizing faces occluded under a surgical mask and skull cap.
Beyond mere 2D human keypoint detection, a significant emphasis has also been placed on regressing keypoints from multiple camera views in a shared 3D space~\cite{liu2022recent,voxelpose}. 
3D human pose detection can be especially beneficial for multi-person scenarios such as surgical ORs, where ubiquitous occlusions can lead to poor performance in individual camera views~\cite{hu2022multi,ozsoy20224d}.
Regressing a 3D human shape from a single input image is also an active area of research~\cite{kolotouros2019convolutional}. 
However, such methods would suffer similarly to partial occlusions. 
To avoid this shortcoming, we leverage the 3D nature of \textit{multi-view} OR acquisitions.

\section{Methods}
\begin{figure}[!hbt]
    \includegraphics[width=\columnwidth]{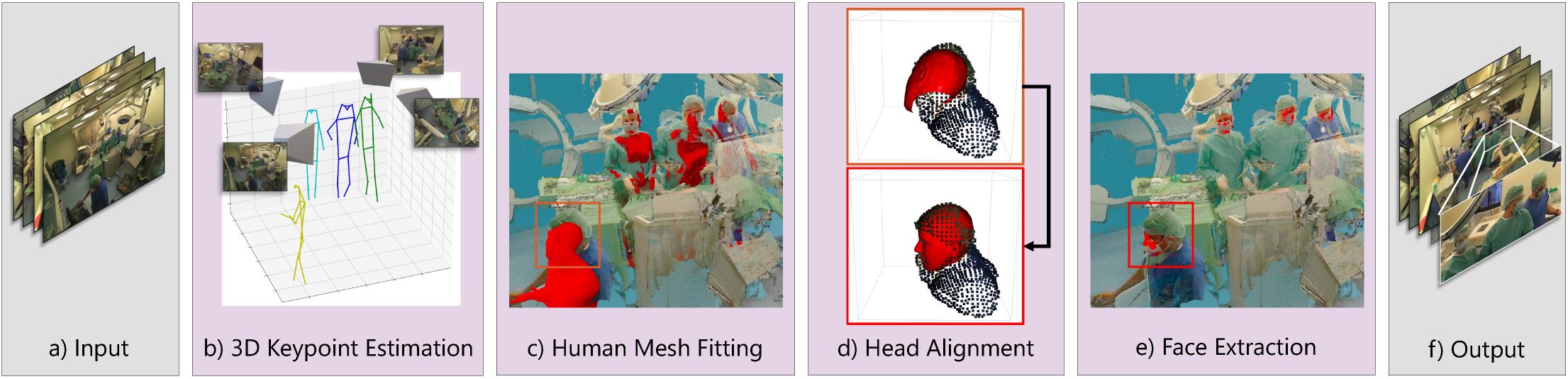}
    \caption[Overview of Our Pipeline]{Anonymization Pipeline of DisguisOR.
    Four RGB images (a) and a fused 3D point cloud as input. Human pose keypoints are first detected in individual views and fused in 3D for each person in the scene (b). Fitting of parametric human mesh model onto keypoints of each person in the scene (c). Further geometrical refinement of the mesh positioning (d) and face extraction (e). 3D mesh texturization and back-projection to each camera view (f)}
    \label{fig:overview}
\end{figure}

An overview of our proposed method DisguisOR is shown in~\autoref{fig:overview}.
We use the \textit{multi-view} OR dataset introduced in Bastian et al.~\cite{know_your_sensors} depicting veterinary laparoscopic procedures, expanded to all four cameras available in the acquisitions. 
Each camera's color and depth images are combined into a colored 3D point cloud using the Azure Kinect framework. 
Subsequently, the four partial point clouds are registered into one global coordinate space by minimizing the photometric reprojection error over keypoints on a large visual marker. 
Our pipeline thus uses RGB images and depth maps from each camera, along with a fused point cloud of the entire scene as input.

\noindent\textbf{Multi-person 3D Mesh Regression.}
We adopt an unsupervised 3-stage approach to fit a 3D mesh~\cite{smpl} for each person in the scene.
2D human keypoints are first detected~\cite{dekr} in each camera view, and regressed in a global coordinate frame with VoxelPose~\cite{voxelpose}.
As neither 2D nor 3D human poses are available as ground truth, we use an existing detector~\cite{dekr} trained on COCO to estimate 2D human keypoints from an image.
In order to combine poses from each view in a robust manner, Voxelpose must first be trained to learn how multiple 2D poses from each camera can be optimally combined in 3D.
To achieve this, we follow the procedure described in~\cite{voxelpose} and synthetically generate ground truth by sampling existing 3D human poses from the Panoptic dataset~\cite{panoptic} and placing them at random locations in the 3D space.
These poses are then projected back into each 2D image-plane and used as input to guide VoxelPose through the 2D-to-3D multi-person regression task.
The trained model ultimately combines 2D human poses from multiple views into one joint 3D human pose for each person in the scene.
We then perform an additional temporal smoothing on each 3D human-pose sequence to interpolate missing poses and reduce noise (for details, see suppl.).

\noindent\textbf{3D Human Representation.}
In order to adequately represent the face of each individual in the scene, we propose to use the statistical parametric human mesh model SMPL~\cite{smpl}, which we regress onto each 3D human pose obtained as output from Voxelpose.
While temporal smoothing yielded less noisy keypoint estimates, we noticed that the 3D mesh model did not always align with the 3D point cloud of an individual, resulting in inaccurate face localizations.
To resolve this issue, we perform a rigid registration between the head segment of the SMPL model and the point cloud.
More specifically, we crop the point cloud around the estimated head of the SMPL model and align the model with the point cloud using the probabilistic point-set registration method FilterReg~\cite{filterreg}, and subsequently fine-tune using iterative closest point (ICP)~\cite{icp}.
As a final step, we extract the face from the SMPL mesh, which should now be aligned with the 3D location of an individual's face.

\noindent\textbf{Rendering the faces in 2D.}
Thus far, our pipeline estimates a global 3D face mesh that overlaps with each person in the scene. In order to yield a 2D anonymization, these meshes can now be projected back into all camera views, replacing the face of each individual with a unique template (see \autoref{fig:naive_anonymization_examples}). 
However, a 3D face might be occluded in a particular view, for example, due to an OR light and, therefore, not visible. 
To mitigate false-positive predictions, we check whether a 3D face is visible in 2D by looking for a disparity between the camera's depth map and the 3D face mesh (for details, see suppl.).
We then utilize the Poisson Image Editing technique~\cite{poisson_image_editing} to harmonize the face template and the background image for a more natural appearing face replacement.
The template can also be changed for each individual to influence factors such as age, sex, or ethnicity.

\subsection{Ground Truth Curation}
\label{subsec:ground_truth_curation}
\begin{figure}[htpb]
    \includegraphics[width=\columnwidth]{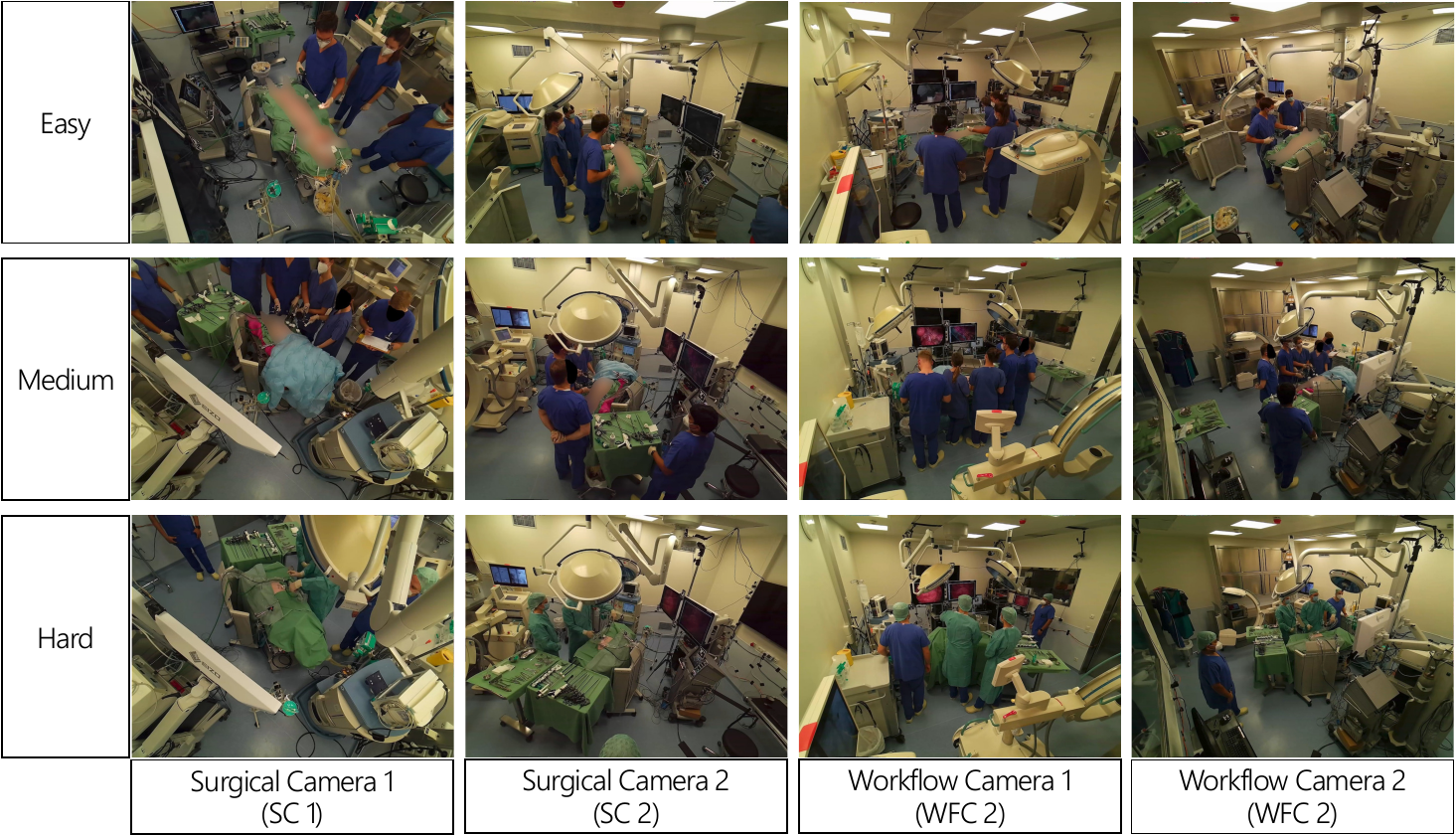}
    \caption{
        Overview of the Three Scenarios and All Four Camera Views.
        We categorize the scenes by their complexity. Surgical cameras (SC) are characterized by persistent obstructions and unusual viewing angles. Workflow cameras (WFC) exhibit ordinary viewing angles with fewer obstructions}
    \label{fig:scenario_overview}
\end{figure}

We identified three distinct scenarios of varying difficulties from the complete dataset~\cite{know_your_sensors}.
We then annotated each visible face manually in all camera views for a total of 4913 face bounding boxes. 
The annotation criteria were adopted to closely match the style of the WIDERFACE dataset~\cite{widerface}.
The three scenarios are chosen to specifically represent the varying characteristics in the OR. They differ in the number of individuals present, their attire, and the degree of obstructions~(\autoref{fig:scenario_overview}).

\begin{itemize}
    \item \textbf{Easy Evaluation Scenario.} %
    Up to four people in the scene, all wearing surgical masks and hospital scrubs with only a few face obstructions. A total of 1310 faces.
    \item \textbf{Medium Evaluation Scenario.} 
    Five or six people in the scene with regular face obstructions caused by the position of the surgical lights. A total of 2317 faces.
    \item \textbf{Hard Evaluation Scenario.} Four people are present in the room. %
    Clinicians additionally wear skull caps and gowns. The surgical lights frequently obstruct the faces in two of the views. A total of 1286 faces.
\end{itemize}

\section{Experiments}
\label{subsec:exp_setup_face_detection}
\noindent\textbf{Face Localization.}
We compare the proposed method's face localization performance with that of DSFD~\cite{dsfd}, a state-of-the-art detector also used in DeepPrivacy~\cite{deepprivacy}. We use the model pre-trained on WIDERFACE~\cite{widerface} provided by the authors. We additionally evaluate the self-supervised domain adaption (SSDA) strategy proposed by Issenhuth et al.~\cite{issenhuth_face_detection_2018}. Here we also use DSFD as the face detection backbone, fine-tuning it on 20k unlabeled images as proposed, with the suggested hyperparameters.

In addition to recall, we propose to evaluate \textit{multi-view} OR anonymization with what we coin \textit{holistic recall}. 
The \textit{holistic recall} considers a face as detected only if it was identified in all camera views where it is at least partially visible.
We argue that this is more suitable than image-wise evaluation, as a missed detection of a face in a single view results in a breach of anonymization for that individual.

We calculate the smallest rectangle outlining the rendered mesh to generate face predictions for evaluation. 
As the proposed method does not rank the output detections with a confidence score, the commonly used average precision (AP) score is not defined. Therefore, we additionally report precision and F1-score for all three methods in the supplementary materials.
Furthermore, the four cameras are categorized as either a \textit{surgical camera} (SC) or \textit{workflow camera} (WFC), depending on the perspective of the camera (see \autoref{fig:scenario_overview}).
The images and angle of acquisition in WFCs are more similar to what might be found in public face detection datasets~\cite{widerface}, while SCs may acquire the scene from above, and individuals are more frequently obscured by OR equipment.

\label{subsec:anonymization}
\noindent\textbf{Image Quality.} We compare the images anonymized by our approach to those altered by several conventional anonymization methods, such as blurring (61x61 kernel), pixelization (8x8 pixels), blackening, as well as the established GAN-based model DeepPrivacy~\cite{deepprivacy} (see 2D anonymization~\autoref{fig:teaser_figure}).
To disentangle image quality and face detection, we only evaluate image quality on faces detected by both our method and Deep Privacy, totaling 3786 faces.
We evaluate the effectiveness of our face replacements on the cropped ground-truth bounding boxes with three established image quality metrics.
The fréchet inception distance (FID)~\cite{fid} measures the overall realism by calculating the distribution distance of the original and generated set of images.
Learned perceptual image patch similarity (LPIPS)~\cite{lpips} reflects the human perception of an image's realism by computing the difference between activations of two image patches for a standard neural network.
The structural similarity index measure (SSIM)~\cite{ssim} calculates the quality of an image pixel-wise based on luminance, contrast, and structure.

Finally, we conduct additional experiments on the downstream behavior of off-the-shelf methods on our anonymized faces (see suppl.)

\section{Results}
\subsection{Face Detection}
\begin{figure}
    \includegraphics[width=\columnwidth]{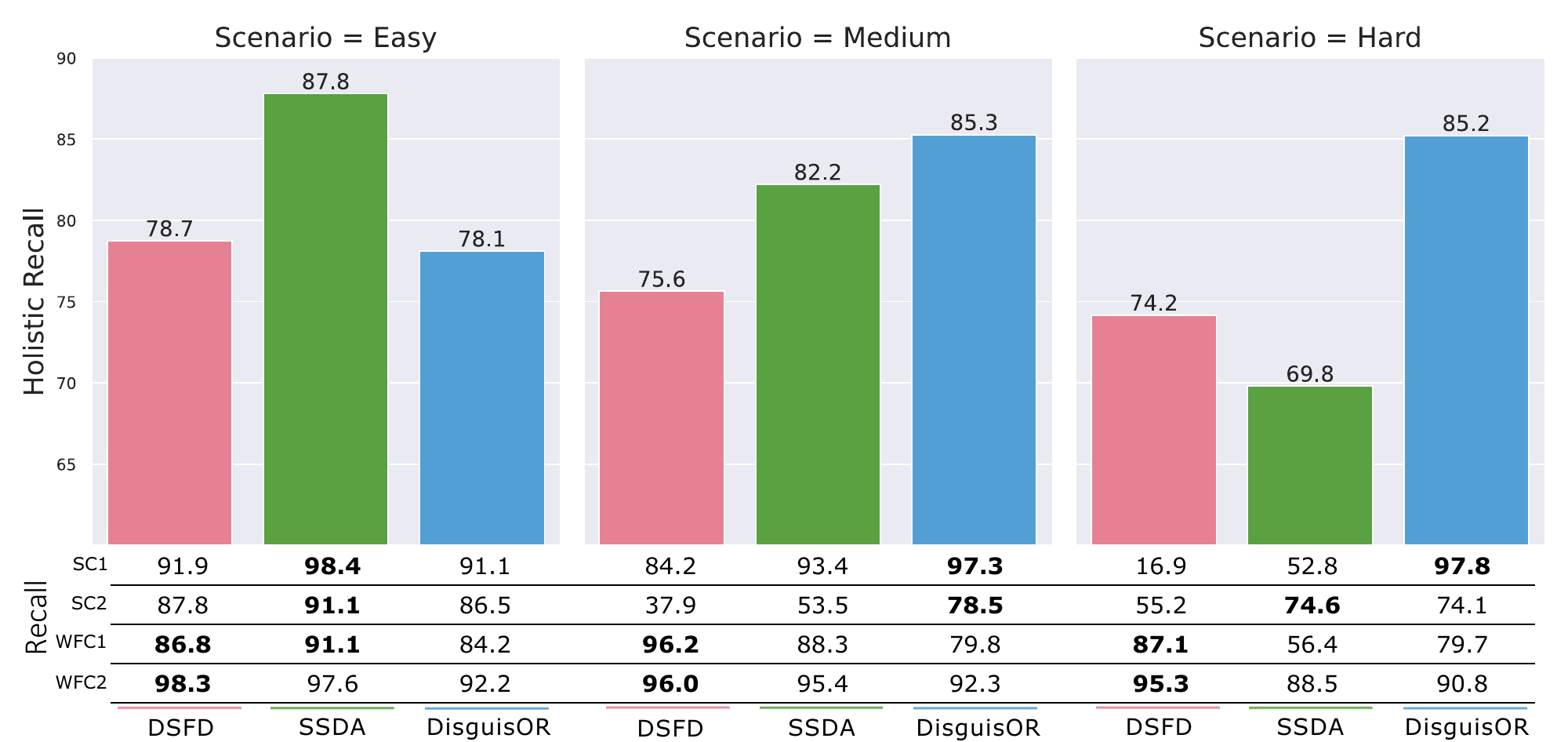}
    \caption{
        Face Localization Performance of DSFD~\cite{dsfd}, the Self-Supervised Domain Adaption (SSDA) Method of~\cite{issenhuth_face_detection_2018}, and DisguisOR Over All Scenarios.
        The \textit{holistic recall} considers a face as detected only if the same face was successfully detected in all camera views where it is at least partially visible. Both recall and \textit{holistic recall} are reported for IOU@0.4
        }
    \label{fig:holistic_recall}
\end{figure}

\autoref{fig:holistic_recall} depicts the performance of our proposed method in comparison with two existing baselines.
In the easy evaluation scenario, both DSFD and DisguisOR perform comparably, while the SSDA achieves a 9\% higher \textit{holistic recall}. 
In the medium and hard scenarios, DisguisOR outperforms DSFD and SSDA in \textit{holistic recall} by 10\% and 3\%, and 11\% and 16\%, respectively.\\
These disparities are largely due to a poor detection rate in the surgical cameras, which are acquired from unusual camera angles and contain frequent obstructions (\autoref{fig:scenario_overview}).
DisguisOR is able to better cope with the increased occlusions and the number of individuals present in these scenarios, highlighting the proposed method's robustness under partial visibility.
By combining information from multiple cameras, DisguisOR yields a geometrically consistent detection -- if an individual face has been accurately localized in the 3D scene, it can be more consistently identified in each individual image.
While DSFD achieves a significantly higher accuracy in the easy scenario when refined via SSDA, the human pose backbone of DigsuisOR underwent no such refinement and would likely also see some performance improvements.

SSDA underperforms the baseline DSFD model in the hard scenario, as well as DisguisOR in both the medium and hard scenarios. 
This could be because these more challenging detection candidates are less represented in the training data distribution or not detected with a high confidence score, and thus not pseudo-labeled frequently enough.

The recall rates over individual cameras reflect the characterizations of surgical and workflow camera views. While DSFD generally achieves slightly higher recall rates on workflow camera views (WFC1, WFC2), DisguisOR achieves much higher recall rates in the surgical camera views (SC1, SC2), see~\autoref{fig:scenario_overview}.
SSDA improves recall for DSFD in surgical cameras, although it still falls short of DisguisOR in medium and hard scenarios.
The surgical cameras in the hard scenario are especially challenging for face detectors, as severe occlusions, unusual camera angles, and surgical scrubs drastically impair the face detectors' detection rate.
In the case of faces in SC1 of the hard scenario (see person 1 in~\autoref{fig:teaser_figure}), DSFD achieves a recall rate of 16.9\%. Using SSDA increases this recall rate to 52.8\%, which DisguisOR still outperforms with a recall rate of 97.8\%.

Our method is somewhat limited by the field-of-view (FOV) of the depth sensors during an acquisition. 
This partially explains the comparable performance with DSFD in the easy scenario, as individuals frequently move along the edge of the scene where depth coverage is limited.
The 3D reconstruction we use to triangulate faces could also be performed without the use of the slightly more costly depth sensors, albeit less accurately.

\subsection{Image Quality}
\begin{figure}[htpb]
    \includegraphics[width=\columnwidth]{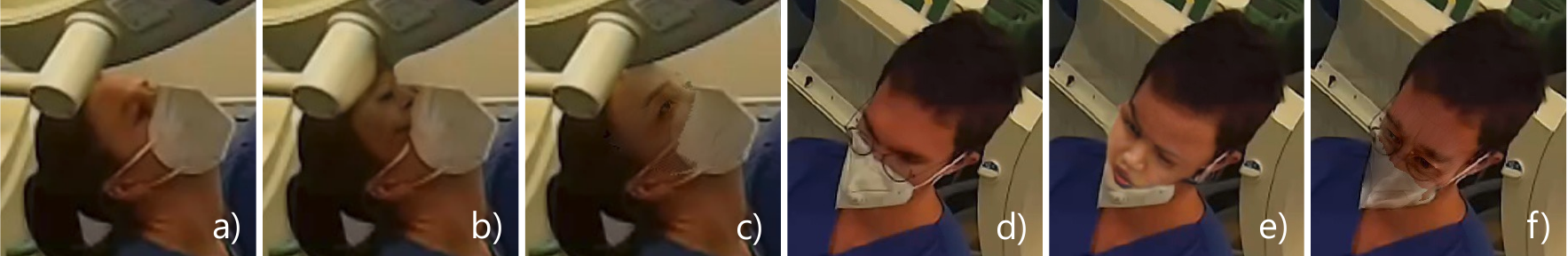}
    \caption{
        Faces From Our Dataset Anonymized With Different Methods.
        Two unaltered faces (a, d), faces anonymized with DeepPrivacy~\cite{deepprivacy} (b, e) and anonymized with DisguisOR (c, f). Note that DeepPrivacy fails to incorporate face masks in its generated faces. Rendering faces with a texture that reflects the setting (in this case a texture with a mask) can yield more consistent face replacement results}
    \label{fig:naive_anonymization_examples}
\end{figure}

In \autoref{tbl:comparison_image_quality_metrics}, we measure the quality of images altered by baseline approaches and our proposed method. 
As expected, conventional obfuscations like blackening, pixelization, and blurring achieve inferior results across all three metrics. 
DeepPrivacy~\cite{deepprivacy} is designed to generate synthetic faces instead of applying conventional privacy filters, explaining the improved results on all image quality metrics compared to the conventional methods. 
Our method further improves upon these results as the replacement of the face information can be precisely controlled, even enabling the replacement of people wearing masks without creating corrupted or unnatural faces.
In \autoref{fig:naive_anonymization_examples}, we illustrate examples where DeepPrivacy replaces the face mask of a person with an unnatural mouth (e), while our method manages to blend the template and original image (f) more effectively.

\begin{table}
    \centering
    \caption[Image Quality Metrics of Us, DeepPrivacy and Conventional Methods]{Comparison of different anonymization techniques based on image quality metrics. An arrow depicts whether a smaller (down) or larger (up) value is more favorable with respect to each metric
    \label{tbl:comparison_image_quality_metrics}}
    \begin{tabular}{l|ccc}
        \toprule
        Method & FID ↓~\cite{fid} & LPIPS ↓~\cite{lpips} & SSIM ↑~\cite{ssim}\\ \midrule
        Blackening & 194.03 & 0.5392 & 0.1864\\
        Pixel & 173.34 & 0.4037 & 0.6080\\
        Blur & 164.57 & 0.3688 & 0.6014\\
        DeepPrivacy~\cite{deepprivacy} & 94.85  & 0.2276 & 0.6294\\
        DisguisOR & \textbf{35.24} & \textbf{0.1341} & \textbf{0.8143}\\
        \bottomrule
    \end{tabular}
\end{table}

\section{Conclusion}
Existing anonymization methods do not effectively leverage \textit{multi-view} data as they consider individual views independently.
OR cameras are frequently mounted in unconventional positions and therefore suffer from heavy occlusions, making multiple views essential for accurately acquiring details of a procedure.
Our 3D face detection framework DisguisOR enables consistency over each camera, preventing missed detections in a single view that would breach an individual's anonymity.
Therefore, we emphasize the use of scene-level anonymization with our proposed \textit{holistic recall} metric to consider the recall of faces detected jointly in all camera views.
We validate our face detection approach based on recall on individual camera views as well as \textit{holistic recall}, demonstrating that our method achieves state-of-the-art results under challenging scenarios and views.

Furthermore, anonymization methods must balance the discrepancy between anonymizing data and retaining its downstream utility.
We show that our framework reduces this discrepancy by yielding more realistic face replacements compared to existing methods.
The modularity of our anonymization approach provides us with fine-grained control of the face replacement, allowing us to vary parameters such as age, gender, or ethnicity.
Existing datasets could even be augmented with faces representing a broad demographic, combating bias induced by unrepresentative training sets.
We are convinced that our method will facilitate further research by reducing the burden of manually annotating existing and future \textit{multi-view} data acquisitions.\\

\noindent\textbf{Declarations, Acknowledgements, and Ethical Compliance.}\\
No ethical approval was required for this study. Consent to use un-anonymized content in figures was obtained by the participants.
This work was funded by the German Federal Ministry of Education and Research (BMBF), No.: 16SV8088 and 13GW0236B. We additionally thank the J\&J Robotics \& Digital Solutions team for their support.
Code will be made available at: https://github.com/wngTn/disguisor\\

\bibliography{bibliography} %

\end{document}


\author{\fnm{Lennart} \sur{Bastian$^*$}}
\author{\fnm{Tony Danjun} \sur{Wang$^*$}}
\author{\fnm{Tobias} \sur{Czempiel}}
\author{\fnm{Benjamin} \sur{Busam}}
\author{\fnm{Nassir} \sur{Navab}}

\affil{\orgdiv{Computer Aided Medical Procedures}, \orgname{Technical University Munich}, \orgaddress{\city{Munich},  \country{Germany}}}
\affil{\small{\\$^*$ Equal Contribution. \qquad Contact: \{first\}.\{last\}@tum.de}}

\title[DisguisOR]{DisguisOR -- Supplementary Materials}

\maketitle

\section{Methods}

\subsection{Obstruction Detection}
\begin{figure}[ht]
    \centering
    \includegraphics[width=.95\columnwidth]{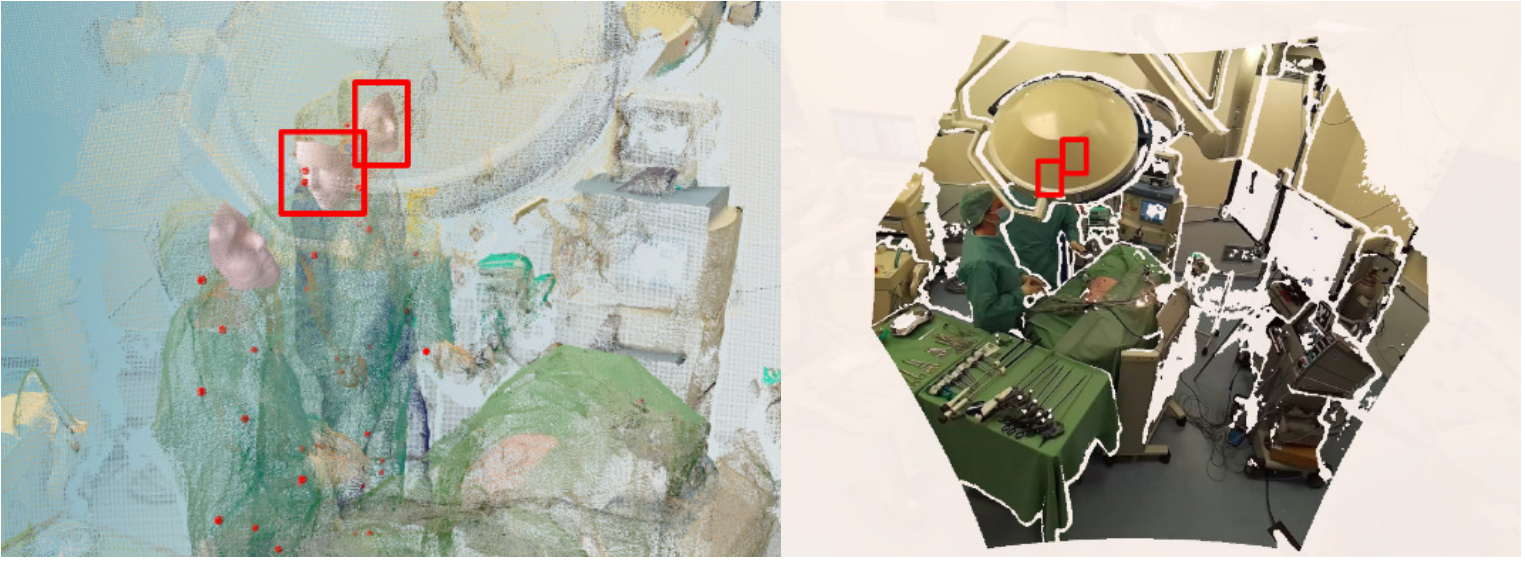}
    \caption[Objects that are obstructed in a specific view]{
        Obstructed faces in the pointcloud and color image.
        The left image shows three face meshes in 3D, the right image shows the depth map of surgical camera 2.
        The two red rectangles reveal the positions of two face meshes in both images.
        Our method checks for obstructions by calculating the distance between the meshes' 3D locations and the inferred 3D locations from the depth map
        }
    \label{fig:reprojection}
\end{figure}

In~\autoref{fig:reprojection}, we detail how we handle obstructions that occur in the back-projection of the 3D face mesh. If an object is positioned between the face and a camera then the calculated distance between the camera and face in 3D will be larger than the distance inferred from the depth.
Specifically, we obtain the 2D coordinates of 20 fixed vertices of the 3D face mesh for each camera view. 
Using the depth map, we calculate the corresponding 3D coordinates using the depth values of each 2D coordinate. 
If the distance between a 3D coordinate inferred from the depth map and the corresponding actual 3D coordinate of a vertex is similar, we conclude the face is visible for the camera in question.

\begin{figure}[htpb]
    \includegraphics[width=\columnwidth]{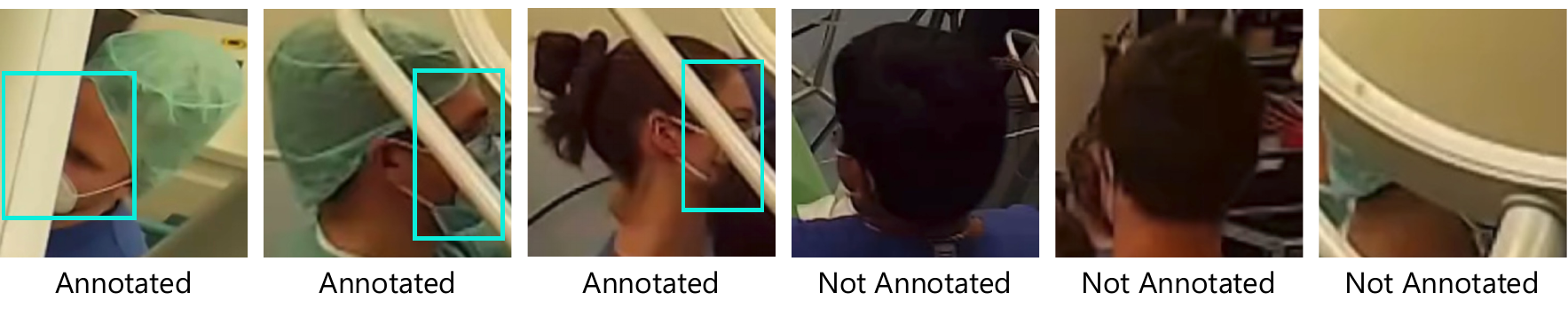}
    \caption{
        Examples of faces from our dataset with their respective annotation.}
    \label{fig:annotation_examples}
\end{figure}

\subsection{Dataset Curation}
In~\autoref{fig:annotation_examples}, we provide a set of examples of the ground truth annotations in our dataset. It denotes the annotation bounding boxes of obstructed faces and cases where no face information is visible.

\subsection{3D Key-Point Smoothing}
To generate less noisy and more robust keypoints we perform a 3D keypoint smoothing and tracking over an entire sequence before fitting the SMPL mesh onto the 3D keypoints. 
The smoothing process helps to diminish outliers caused by incorrect triangulations during the 3D human pose estimation stage and to interpolate missed human poses.
This allows for a more reliable human mesh regression onto the 3D human poses.
Furthermore, the tracking enables anonymizing each person with the same face texture in every frame. 

\subsection{Rendering: Poisson Image Editing}
\begin{figure}
    \includegraphics[width=\columnwidth]{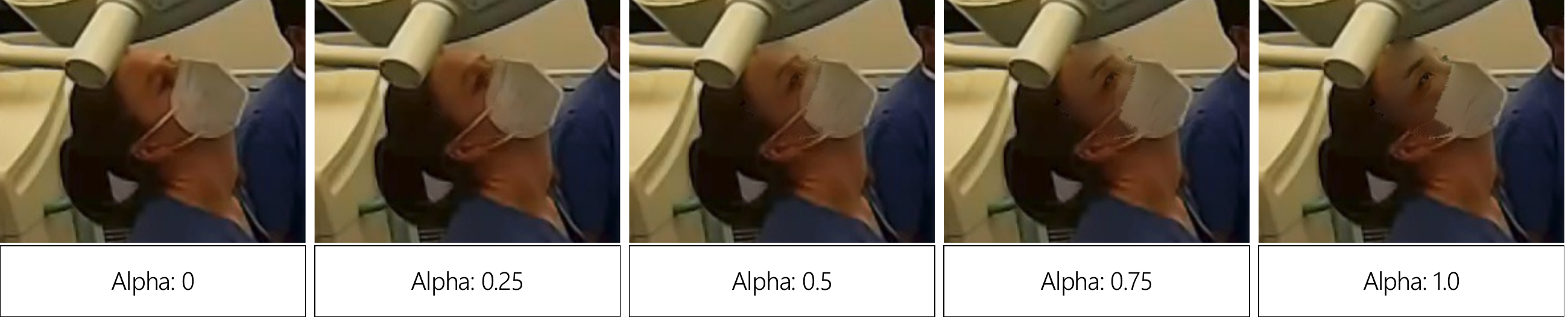}
    \caption{
        Examples of tuning the parameter for image blending with poisson image editing. With an alpha value of 1.0, no gradient of the background image is used, while a value of 0 uses only the gradients of the background image. We chose 0.725 as parameter to balance image quality and sufficient anonymization
    }
    \label{fig:alpha_scores}
\end{figure}

\begin{figure}
    \centering
    \includegraphics[width=\columnwidth]{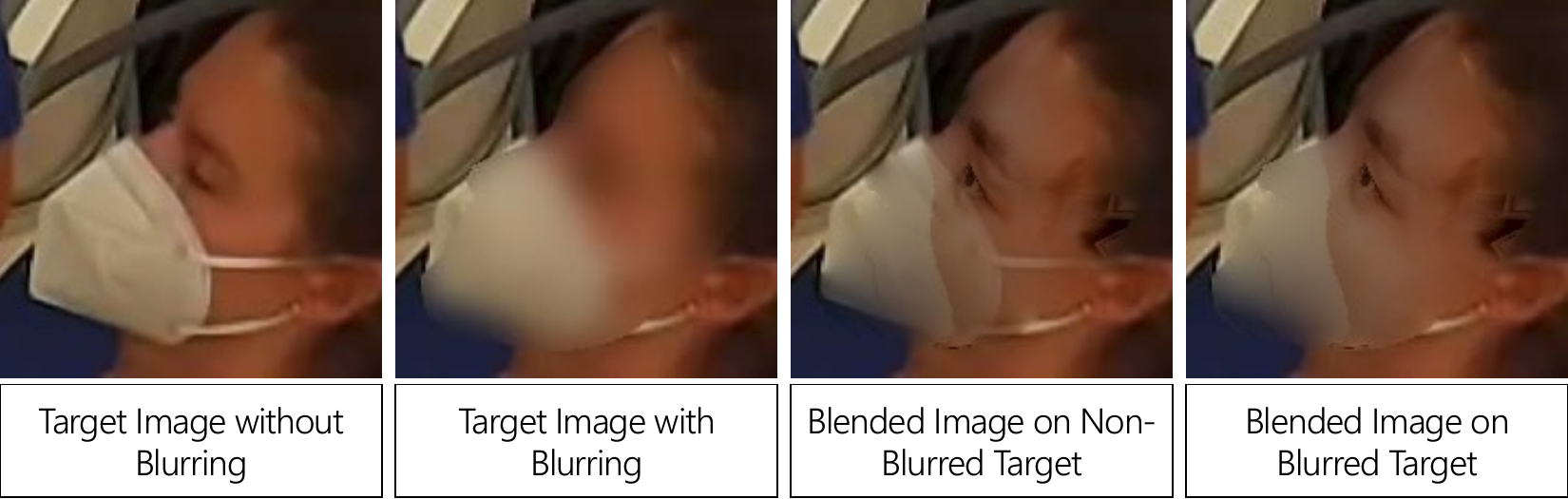}
    \caption{
        Illustration of combining poisson image editing with blurring. Blurring the target image allows to further hide prominent background information in the result. The medical mask's contour of the original target blends seamlessly with the final output
    }
    \label{fig:poisson_blurring}
\end{figure}

We further highlight the effect of different blending parameters of the poisson image editing image harmonization~(\autoref{fig:alpha_scores}).
As an additional robustness towards privacy preservation, one may also opt to blur in the source image prior to blending~(\autoref{fig:poisson_blurring}) or avoid blending altogether ~(\autoref{fig:alpha_scores} Alpha: 1.0).

\section{Results and Discussion}

\subsection{Face Localization}
In Tables~\ref{tbl:easy_eval_dataset_results},~\ref{tbl:medium_eval_dataset_results} and~\ref{tbl:hard_eval_dataset_results}, we present the precision (P), recall (R), F1-score (F1), and number of annotated faces of each individual camera and scenario.
The precision of DisguisOR is lower for some scenarios than that of DSFD, notably in the surgical cameras (SC1 and SC2). 
This is mainly due to the hardware constraints of the Kinect camera system, which does not generate a corresponding depth value for each pixel.
Cropping to the depth field of view and downsizing the color image to the resolution of the depth camera would mitigate these issues and likely favor DisguisOR -- however, we perform inference for all methods at native resolution of 2048x1536.
While the self-supervised domain adapation (SSDA)~\cite{issenhuth_face_detection_2018} with DSFD increased the recall significantly, especially in the surgical cameras, it also decreased the precision, resulting in a lower F1-score.

\begin{table}[h]
\centering
    \caption[]{Comparison of precision (P), recall (R) and F1-score (F1) at IOU@0.4 of DSFD~\cite{dsfd}, the Self-Supervised Domain Adaption (SSDA) method of~\cite{issenhuth_face_detection_2018}, and DisguisOR on the \textbf{Easy Scenario} across all cameras. The last column depicts the number of ground truth faces in each camera view
        \label{tbl:easy_eval_dataset_results}}
        \begin{tabular*}{\textwidth}{@{\extracolsep{\fill}}lccc|ccc|ccc|c}
            \hline
            \  & \multicolumn{3}{c|}{DSFD~\cite{dsfd}} & \multicolumn{3}{c|}{SSDA~\cite{issenhuth_face_detection_2018}} & \multicolumn{3}{c|}{DisguisOR} & \multicolumn{1}{c}{No. Faces}\\
                 & P & R & F1 & P & R & F1 & P & R & F1 &  \\ \hline
            SC1  & 90.5 & 91.9 & 91.2 & 44.8 & 98.4 & 61.6 & 83.5 & 91.1 & 87.1 & 384 \\ \hline
            SC2  & 87.2 & 87.8 & 87.5 & 71.6 & 91.1 & 80.2 & 75.7 & 86.5 & 80.7 & 327 \\ \hline
            WFC1 & 69.3 & 86.8 & 77.1 & 51.0 & 91.1 & 65.4 & 61.8 & 84.2 & 71.3 & 190 \\ \hline
            WFC2 & 97.8 & 98.3 & 98.1 & 62.2 & 97.6 & 76.0 & 87.3 & 92.2 & 89.7 & 409 \\ \hline
            Avg. & 86.2 & 91.2 & 88.5 & 57.4 & 94.6 & 70.8 & 77.1 & 88.5 & 82.2 & 327 \\ \hline
        \end{tabular*}
\end{table}

\begin{table}[h]
\centering
    \caption[]{Comparison of precision (P), recall (R) and F1-score (F1) at IOU@0.4 of DSFD~\cite{dsfd}, the Self-Supervised Domain Adaption (SSDA) method of~\cite{issenhuth_face_detection_2018}, and DisguisOR on the \textbf{Medium Scenario} across all cameras. The last column depicts the number of ground truth faces in each camera view
        \label{tbl:medium_eval_dataset_results}}
        \begin{tabular*}{\textwidth}{@{\extracolsep{\fill}}lccc|ccc|ccc|c}
            \hline
            \  & \multicolumn{3}{c|}{DSFD~\cite{dsfd}} & \multicolumn{3}{c|}{SSDA~\cite{issenhuth_face_detection_2018}} & \multicolumn{3}{c|}{DisguisOR} & \multicolumn{1}{c}{No. Faces} \\
                 & P & R & F1 & P & R & F1 & P & R & F1 &  \\ \hline
            SC1  & 97.5 & 84.2 & 90.4 & 71.9 & 93.4 & 81.3 & 87.4 & 97.3 & 92.1 & 558  \\ \hline
            SC2  & 63.4 & 37.9 & 47.4 & 29.6 & 53.5 & 38.1 & 35.1 & 78.5 & 48.5 & 256  \\ \hline
            WFC1 & 87.2 & 96.2 & 91.5 & 67.1 & 88.3 & 76.3 & 71.7 & 79.8 & 75.5 & 426  \\ \hline
            WFC2 & 94.0 & 96.0 & 95.0 & 94.0 & 95.4 & 94.7 & 93.3 & 92.3 & 92.8 & 1077 \\ \hline
            Avg. & 85.5 & 78.6 & 81.1 & 65.7 & 82.7 & 72.6 & 71.9 & 87.0 & 77.2 & 579  \\ \hline
        \end{tabular*}
\end{table}

\begin{table}[h]
\centering
    \caption[]{Comparison of precision (P), recall (R) and F1-score (F1) at IOU@0.4 of DSFD~\cite{dsfd}, the Self-Supervised Domain Adaption (SSDA) method of~\cite{issenhuth_face_detection_2018}, and DisguisOR on the \textbf{Hard Scenario} across all cameras. The last column depicts the number of ground truth faces in each camera view
        \label{tbl:hard_eval_dataset_results}}
        \begin{tabular*}{\textwidth}{@{\extracolsep{\fill}}lccc|ccc|ccc|c}
            \hline
            \  & \multicolumn{3}{c|}{DSFD~\cite{dsfd}} & \multicolumn{3}{c|}{SSDA~\cite{issenhuth_face_detection_2018}} & \multicolumn{3}{c|}{DisguisOR} & \multicolumn{1}{c}{No. Faces} \\
                 & P & R & F1 & P & R & F1 & P & R & F1 &  \\ \hline
            SC1  & 88.2 & 16.9 & 28.4 & 10.3 & 52.8 & 17.2 & 19.6 & 97.8 & 32.7 & 89  \\ \hline
            SC2  & 88.3 & 55.2 & 67.9 & 89.6 & 74.6 & 81.4 & 62.3 & 74.1 & 67.7 & 232 \\ \hline
            WFC1 & 76.2 & 87.1 & 81.3 & 73.5 & 56.4 & 63.8 & 51.1 & 79.7 & 62.3 & 202 \\ \hline
            WFC2 & 99.2 & 95.3 & 97.2 & 73.2 & 88.5 & 80.1 & 93.8 & 90.8 & 92.3 & 763 \\ \hline
            Avg. & 88.0 & 63.6 & 68.7 & 61.7 & 68.1 & 60.7 & 56.7 & 85.6 & 63.7 & 321 \\ \hline
        \end{tabular*}
\end{table}

\subsubsection{Different Confidence Thresholds}

\begin{figure}[htpb]
    \includegraphics[width=\columnwidth]{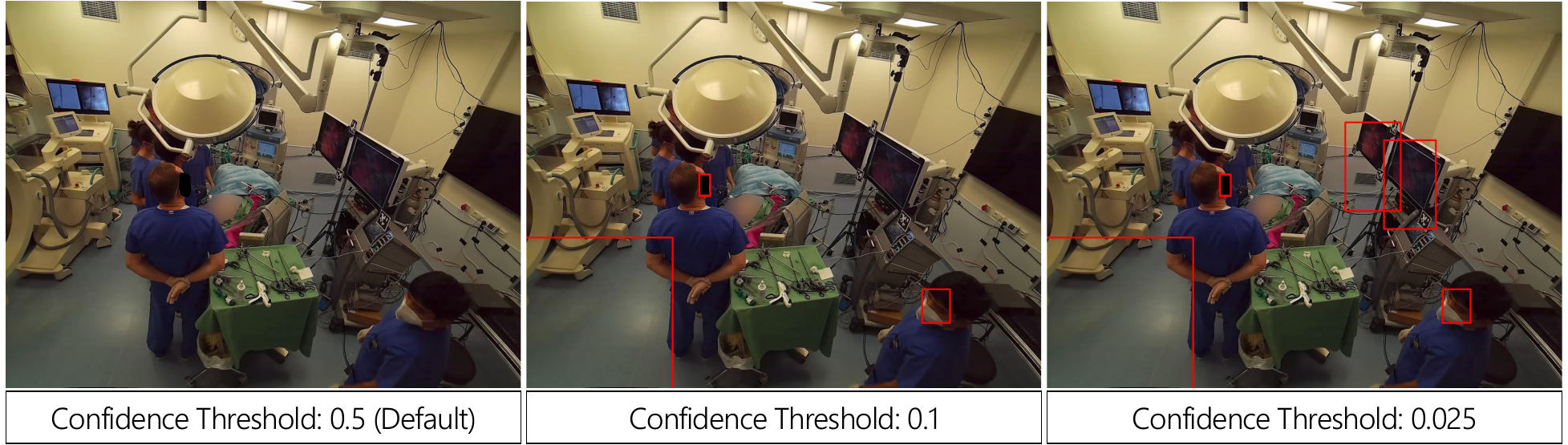}
    \caption{
        Detections of DSFD~\cite{dsfd} with different confidence thresholds. While lowering the threshold of DSFD leads to additional correct predictions, this typically comes at the expense of errant predictions which cover large portions of the image
    }
    \label{fig:conf_threshold}
\end{figure}

\autoref{tbl:conf_thresh_dataset_results} depicts the precision, recall and F1-scores of DSFD~\cite{dsfd} using different confidence thresholds. 
The default confidence threshold is 0.5. 
We average each metric across all four cameras, and present the results for each scenario.
Lowering the confidence scores increases the recall, but at a significant cost of precision. 
Even at lower confidence thresholds, DSFD does not attain the recall of DisguisOR.
Furthermore, errant detections cover large swaths of the images (\autoref{fig:conf_threshold}).
Subsequent anonymization s.pdf could impact the image integrity and affect downstream tasks.

\begin{table}[h]
    \centering
        \caption[]{Comparison of precision (P), recall (R) and F1-score (F1) at IOU@0.4 of DSFD~\cite{dsfd} with different confidence thresholds and DisguisOR on all scenarios. P, R and F1 are averaged across all cameras}
        \label{tbl:conf_thresh_dataset_results}
        \begin{tabular*}{\textwidth}{@{\extracolsep{\fill}}lccc|ccc|ccc} \hline
            \  & \multicolumn{3}{c|}{DSFD~\cite{dsfd} @ 0.1} & \multicolumn{3}{c|}{DSFD~\cite{dsfd} @ 0.025} & \multicolumn{3}{c}{DisguisOR} \\
                  & P & R & F1 & P & R & F1 & P & R & F1  \\ \hline
            Easy    & 73.3 & 94.8 & 81.7 & 12.5 & 96.5 & 20.6 & 77.1 & 88.5 & 82.2 \\ \hline
            Medium  & 71.2 & 84.0 & 76.4 & 26.9 & 88.4 & 38.3 & 71.9 & 87.0 & 77.2 \\ \hline
            Hard    & 64.1 & 69.6 & 64.8 & 23.6 & 73.6 & 28.9 & 56.7 & 85.6 & 63.7 \\ \hline
        \end{tabular*}
\end{table}

\begin{figure}
    \includegraphics[width=\columnwidth]{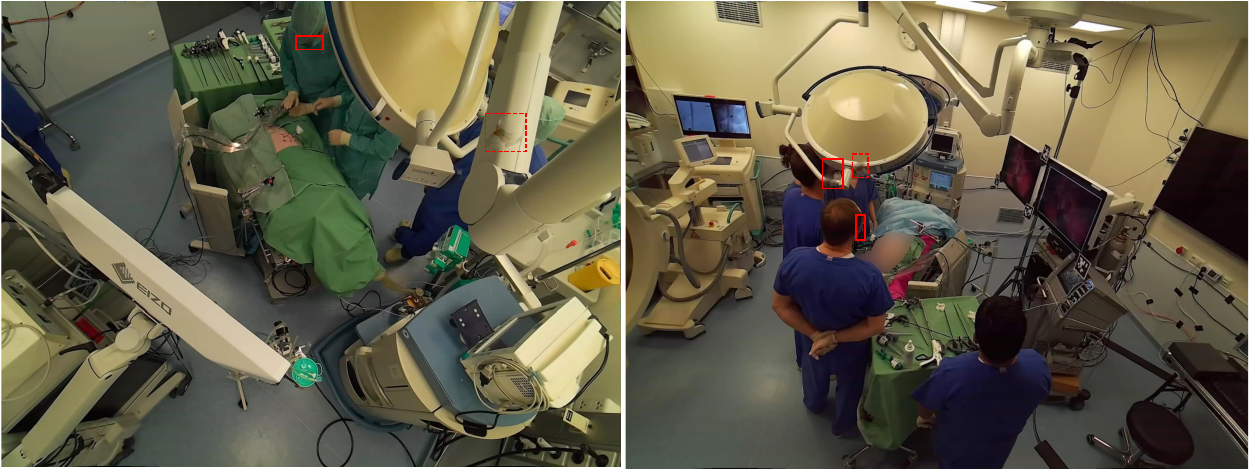}
    \caption{
        Illustration of two different frames, each with a false positive detection due to a failed occlusion check (best viewed digitally). Dotted red face bounding boxes denote false positive detections, while solid bounding boxes depict true positive detections.
        Many DisguisOR false positive predictions are due to failed occlusion checks.
        However, these artifacts represent a small area of the image
    }
    \label{fig:precision_depth}
\end{figure}

\pagebreak
\subsection{Experiments on Downstream Tasks}

\noindent\textbf{Face Detection.}
To evaluate the preservation of image integrity, we compare our anonymization method with the three conventional anonymization methods (blurring, pixelization and blackening) and DeepPrivacy~\cite{deepprivacy} on downstream face detection.
This experiment is designed to measure the degree to which an anonymization method creates unnatural alterations, which would lead to errant predictions from an existing method.
We compare the performance of the pre-trained DSFD~\cite{dsfd} model on images from all three scenarios anonymized through several methods.
For a fair comparison between the two methods, we only consider face detections made by both DeepPrivacy and DisguisOR.
We report the percentage of average precision (AP) at an intersection over union of greater than 0.4 (IOU@0.4) retained with respect to the original unaltered images.

\noindent\textbf{Human Pose Estimation.}
In order to further demonstrate the image quality preservation of our method, we evaluate the widely used AlphaPose~\cite{alphapose} human pose estimator, pre-trained on COCO~\cite{coco_dataset}, to understand the effect of unnatural image artifacts on human pose estimation.
We generate pseudo-ground truth by projecting the 3D joint positions of each generated SMPL mesh into each camera view. 
In accordance with previously established works~\cite{issenhuth_face_detection_2018}, we use the percentage of correct keypoints (PCK) metric to measure how many keypoints were accurately predicted within a threshold.
DeepPrivacy is omitted from this experiment to avoid an unfair bias in favor of DisguisOR due to how pseudo-ground truth is generated.

\section{Results.}

\noindent\textbf{Face Detection.}
\autoref{tbl:effect_anonymization_face_detection} depicts the downstream face detection performance of the face detector DSFD~\cite{dsfd} on images anonymized through various obfuscation methods. 
We report the percentage AP retained compared to the AP achieved on the original unaltered images. 
The results show that blackening the face area removes a considerable amount of information, making it prohibitively difficult for a method to localize a face. Blurring or pixelization is less detrimental to the detection methods' performance.
Nevertheless, the detection results of both methods are severely impacted by the blurring of face regions. 
This becomes even more evident in difficult scenarios. 
For all three conventional anonymization approaches, the performance drop is severe, and the AP for the face detection task is too low for most use cases.
In contrast, our proposed method generates faces that the detection methods can identify with high accuracy.

In the easy scenario, we observe a decrease of merely 2\% in DSFD's face detection AP, highlighting the preservation qualities of our method.
DeepPrivacy is able to retain more AP in the medium and hard scenarios, most likely since it generates synthetic faces with distinct facial features.
Using a maskless face texture further increases the retained AP by enabling face detectors to rely on more facial elements, such as the nose and mouth.
However, this comes at a cost of image quality, as apparent by the experiments in the main paper.
These results indicate that the realistic faces generated by DisguisOR could mitigate costly annotation and retraining due to an inferior anonymization method.
Furthermore, they also indicate that a given template and source image harmonization may be more suitable for a certain application.

\begin{table}[]
    \centering
    \caption[Face Detection on Anonymized Images]{Percentage of face detection AP of DSFD~\cite{dsfd} at IOU@0.4 for differently anonymized images with the AP on original images as baseline. Masked Texture denotes that blended templates contain medical masks (see~\autoref{fig:alpha_scores}), whereas Maskless Texture denotes that blended templates do not contain medical masks
    \label{tbl:effect_anonymization_face_detection}}
    \begin{tabular*}{\textwidth}{@{\extracolsep{\fill}}l|c|c|c}
        \toprule
        Anonymization Method & Easy Scenario & Medium Scenario & Hard Scenario \\
        \midrule
        Blackening & 12.2 & 10.9 & 9.1 \\
        Gaussian Blur & 18.4 & 19.3 & 7.8 \\
        Pixelization & 72.4 & 60.6 & 63.1\\
        DeepPrivacy~\cite{deepprivacy} & 99.1 & \textbf{98.9} & \textbf{94.2} \\
        DisguisOR (Masked Texture) & 98.2 & 85.3 & 74.7 \\
        DisguisOR (Maskless Texture) & \textbf{100.0} & 92.8 & 82.8  \\
        \bottomrule
    \end{tabular*}
\end{table}

\noindent\textbf{Human Pose Estimation.} In~\autoref{tbl:effect_anonymization_pose_estimation} we report the PCK of the human pose estimator AlphaPose~\cite{alphapose} on differently anonymized frames.
Human pose estimators are less susceptible to limited modifications of the face area. Therefore, the deviation of the PCKs is less severe.
Blurring, pixelization, and blackening of the faces can confuse the methods resulting in a decreased performance. 
It is interesting to see that the anonymization limited to the region of the head still has a measurable negative impact on the joint detection of the hip. 
This again highlights the importance of retaining the information and the need to anonymize without severe information loss. 
The PCK on images anonymized by our method is the highest, demonstrating the realism of our method.

\begin{table}[!hbt]
    \centering
    \caption[Human Pose Estimation on Anonymized Images]{PCKh@0.5 results for AlphaPose~\cite{alphapose} on differently anonymized images. Images are taken from all scenarios across all cameras
        \label{tbl:effect_anonymization_pose_estimation}}
        \resizebox{\textwidth}{!}{
    \begin{tabular}{@{\extracolsep{\fill}}l|ccccccc|c}
        \toprule
        Anonymization Method & Head & Shoulder & Elbow & Wrist & Hip & Knee & Ankle & Avg. \\ \midrule
        Blackening & 58.8 & 76.1 & 71.6 & 63.7 & 66.7 & 31.4 & 25.7 & 56.4 \\
        Pixelization & 71.1 & 79.0 & 74.5 & 65.9 & 72.2 & 32.5 & 26.1 & 61.4 \\
        Gaussian Blur & 72.6 & 81.5 & 76.8 & 67.3 & 74.2 & \textbf{34.1} & 26.3 & 63.0 \\
        DisguisOR & \textbf{78.2} & \textbf{82.0} & \textbf{77.4} & \textbf{67.4} & \textbf{76.2} & 33.9 & \textbf{27.0} & \textbf{65.0} \\
        \bottomrule
    \end{tabular}
    }
\end{table}

\section{Appendix}
For reproducibility, we provide the GitHub repository link and python version for each method that we have used in this paper.
All computations were performed on a computer with 64GB of RAM and an NVIDIA GeForce RTX 2080 Ti.
DeepPrivacy needed approximately 2.44s for anonymizing each frame (i.e., 4 images),while DisguisOR needed approximately 6.47s for each frame (see~\autoref{tbl:disguisor_runtime}). The majority of this time is spent on point cloud alignment and rendering, which could be made more efficient. More specifically, it needed 0.52s for 2D human pose estimation, 0.23 for 3D human pose estimation, 0.62s for human mesh estimation and 3.74s for registration and 1.36s rendering.
The memory footprint of DeepPrivacy reached around 4.6GB, while DisguisOR used approximately 6.5GB.

\begin{table}[h!]
    \centering
    \caption{The runtime for each frame (i.e., 4 images) of DisguisOR split into respective stages
    \label{tbl:disguisor_runtime}}
    \begin{tabular}{l|c}
        \toprule
        Stage & Runtime (seconds) \\
        \midrule
        2D Human Pose Estimation & 0.52 \\  
        3D Human Pose Esimtation & 0.23  \\
        Human Mesh Estimation & 0.62 \\
        Point Cloud Registration & 3.74 \\
        Rendering & 1.36 \\
        \midrule
        Total & 6.47 \\
        \bottomrule
    \end{tabular}
\end{table}

\begin{itemize}
    \item DisguisOR  \url{{https://github.com/wngTn/disguisor}} (Python 3.7.2)
    \begin{itemize}
    \item DEKR~\cite{dekr} \url{https://github.com/HRNet/DEKR} (Python 3.7.2)
    \item VoxelPose~\cite{voxelpose} \url{https://github.com/microsoft/voxelpose-pytorch} (Python 3.10.2) 
    \item EasyMocap~\cite{easymocap} \url{https://github.com/zju3dv/EasyMocap} (Python 3.6.2)
    \item SMPL~\cite{smpl} \url{https://smpl.is.tue.mpg.de}
    \end{itemize}
    \item Evaluation code versions
    \begin{itemize}
    \item DSFD~\cite{dsfd} \url{https://github.com/hukkelas/DSFD-Pytorch-Inference} (Python 3.8.13)
    \item DeepPrivacy~\cite{deepprivacy} \url{https://github.com/hukkelas/DeepPrivacy} (Python 3.8.2)
    \item AlphaPose~\cite{alphapose} \url{https://github.com/MVIG-SJTU/AlphaPose} (Python 3.7.2)
    \end{itemize}
\end{itemize}

\bibliography{bibliography} %